\documentclass[conference]{IEEEtran}
\usepackage{cite}
\usepackage{amsmath,amssymb,amsfonts}
\usepackage{algorithmic}
\usepackage{graphicx}
\usepackage{textcomp}
\usepackage{url}
\usepackage{xcolor}
\usepackage{mathrsfs}
\usepackage{subcaption}
\usepackage{bigstrut}

\def\BibTeX{{\rm B\kern-.05em{\sc i\kern-.025em b}\kern-.08em
    T\kern-.1667em\lower.7ex\hbox{E}\kern-.125emX}}
\begin{document}

\title{Learning Neural Volumetric Field for Point Cloud Geometry Compression}

\author{
\IEEEauthorblockN{Yueyu Hu, Yao Wang}
\IEEEauthorblockA{\textit{Department of Electrical and Computer Engineering} \\
New York University, Brooklyn, NY 11201, USA \\
\{yyhu, yw523\}@nyu.edu}}

\maketitle 
\thispagestyle{plain} 

\begin{abstract}
Due to the diverse sparsity, high dimensionality, and large temporal variation of dynamic point clouds, it remains a challenge to design an efficient point cloud compression method. We propose to code the geometry of a given point cloud by learning a neural volumetric field. Instead of representing the entire point cloud using a single overfit network, we divide the entire space into small cubes and represent each non-empty cube by a neural network and an input latent code. The network is shared among all the cubes in a single frame or multiple frames, to exploit the spatial and temporal redundancy. The neural field representation of the point cloud includes the network parameters and all the latent codes, which are generated by using back-propagation over the network parameters and its input. By considering the entropy of the network parameters and the latent codes as well as the distortion between the original and reconstructed cubes in the loss function, we derive a rate-distortion (R-D) optimal representation. Experimental results show that the proposed coding scheme achieves superior R-D performances compared to the octree-based G-PCC, especially when applied to multiple frames of a point cloud video. The code is available at \url{https://github.com/huzi96/NVFPCC/}.
\end{abstract}

\begin{IEEEkeywords}
Point Cloud Compression, Neural Field, Rate-Distortion Optimization
\end{IEEEkeywords}

\section{Introduction}
We are in an era when the 3D visual applications are emerging. Modern 3D visual capturing devices are widely deployed on autonomous cars, mobile phones, drones, \textit{etc}. In these applications, point clouds serve as the raw  representations for processing and analysis of 3D scenes. Due to the computation limitation at the capturing devices, it is common practices to compress and transmit captured point clouds to remote processors and store them for future analysis. On the other hand, in AR/VR applications, 3D contents represented by point clouds need to be compressed and delivered from servers to end users or between end users. 
Both scenarios require efficient compression schemes for point clouds. 

Due to the higher dimensionality and the sparsity in nature, 3D visual data are generally more difficult to process compared to 2D visual data, \textit{i.e.} images and videos. In terms of compression, the challenges coming with point clouds are mainly twofold.

\textbf{Transform Design}. For images and videos, time-frequency transforms like DCT and DWT are shown to perform well for transform coding. Different from images, where the sampling of the signal is on a dense grid, point clouds are sparsely sampled in the 3D space. In terms of voxel grid occupancy, even a dense point cloud is sparse from the signal point of view. Therefore, existing time-frequency transforms cannot be directly applied to point clouds. 
To tackle this problem, machine learning based point cloud coding schemes are developed. One category of learned point cloud coder follows the octree coding structure. Given the already coded parent nodes~\cite{huang2020octsqueeze} and siblings~\cite{que2021voxelcontext,mao2022mipr} in the voxel grids, the probability distribution of the occupancy in the children nodes is predicted, and directly serves entropy coding. The other category adopts the binary voxel grid data structure, where an 3D auto-encoder is adopted to down-sample a point cloud into compact latent representations~\cite{wang2021multiscale}. 
Although these methods achieve impressive rate-distortion performances, they show limitations when adapting to dynamic point cloud coding.

\textbf{Motion Compensation} 
For conventional 2D videos, prediction and motion compensation are shown to be the effective to exploit the temporal redundancy. 
Due to the memory and computational inefficiency, point clouds are usually not represented as voxel grids but rather octrees~\cite{schnabel2006octree}. Such octrees vary from frame to frame and it is difficult to find the correspondences between octree nodes for motion compensation. Therefore, coding of point cloud videos is far more challenging than 2D videos. The Moving Picture Experts Group~(MPEG) has developed two approaches for point cloud coding~\cite{graziosi2020overview}, where the V-PCC adopts a video codec-based method to code point cloud videos. Such methods rely on projections from 3D points to 2D frames during encoding, and reprojection from 2D to 3D during decoding, introducing distortion and complexity. However, because it can leverage the significant progress in 2D video coding over the past 30 years, and effectively exploit the temporal redundancy through motion compensation, V-PCC has substantially better performance than G-PCC applied to individual frames.  
In \cite{fan2022d}, a learned inter-frame predictive point cloud coding framework is developed, and achieves improved R-D performance over V-PCC. However, explicit motion compensation is still required over voxel grids.

\begin{figure*}
    \centering
    \begin{subfigure}[h]{0.60\linewidth}
    \includegraphics[width=1\linewidth]{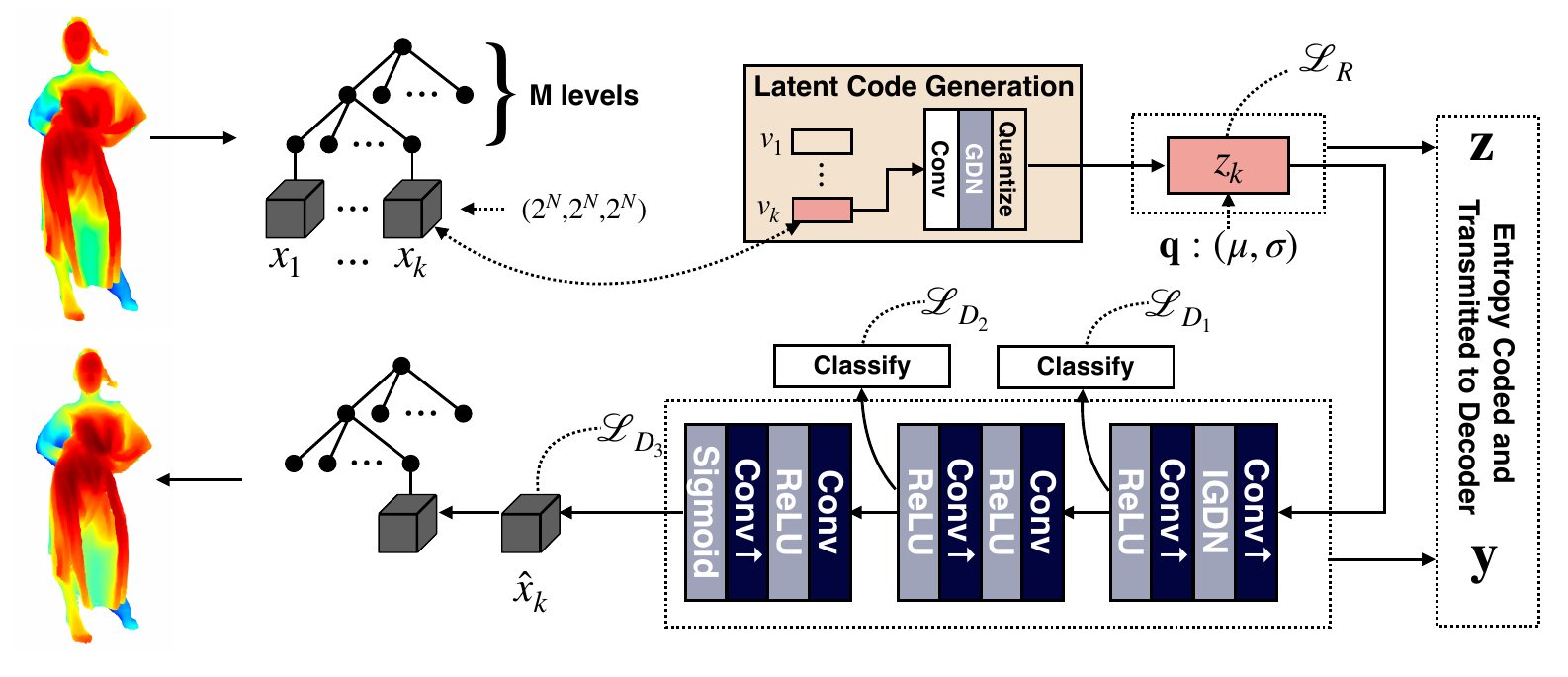}
    \caption{Network structure.}
    \label{fig:structure}
    \end{subfigure}
    \begin{subfigure}[h]{0.3\linewidth}
    \includegraphics[width=1\linewidth]{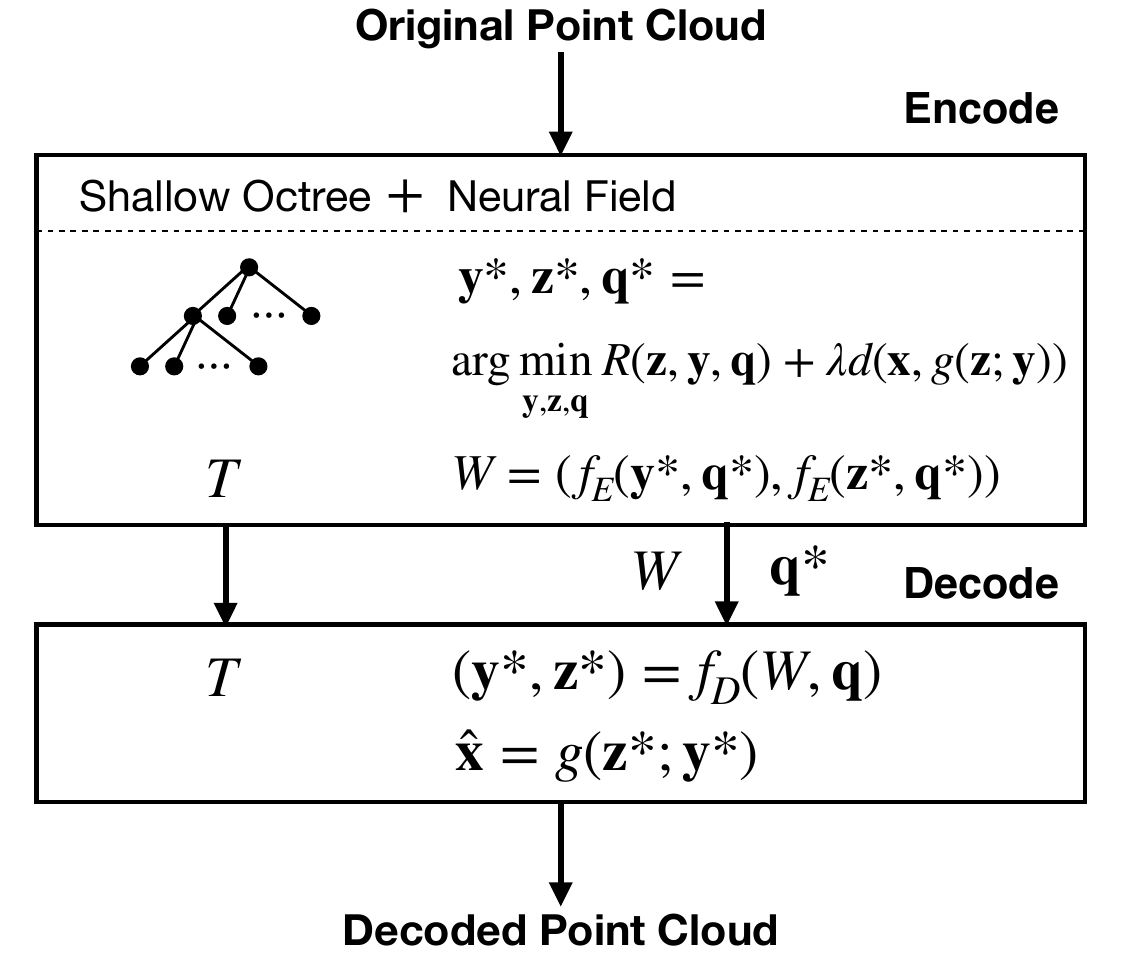}
    \caption{Coding procedure.}
    \end{subfigure}
    
  
    \caption{Overall framework of the proposed method.}
    \vspace{-6mm}
    \label{fig:frame}
\end{figure*}

In this paper, we present a unified approach suitable to compress both static and dynamic point clouds, without explicit motion compensation. 
We focus on the coding of the geometry only.  Our work is inspired by a recent advance in 3D modeling, \textit{i.e.} Neural Radiance Field~(NeRF)~\cite{mildenhall2020nerf}. Our core idea is to represent a  3D point cloud with an implicit neural field. Specifically, we divide the space occupied by a point cloud into subregions based on a shallow octree. Each subregion is associated  with a latent code. The occupancies inside a subregion are reconstructed by its latent code and a network that is shared among all subregions.
Both the network parameters and the latent representations are rate-distortion optimized, quantized and entropy coded. The collection of the network parameters and all the latent codes forms the neural field. To decode, we execute the network with the latent codes and the network parameters to reconstruct the point cloud. This method can be easily extended to code a point cloud video,  by learning a single shared network for all the subregions over a group of frames.

The benefits of our approach are three folds:
\begin{itemize}
    \item We take advantage of machine learning to design non-linear transforms suitable for point cloud signals. Such transforms are effective to reduce the redundancy.
    \item The same method works on both static and dynamics point clouds, yielding more compression gains  on dynamic point clouds without explicit motion estimation.
    \item Our approach does not rely on any dataset to train, providing more flexibility for the highly diverse point cloud data.
\end{itemize}
%
%
 Experimental results show that the proposed method achieves superior rate-distortion performance compared with G-PCC\cite{gpcc}, the MPEG octree based point cloud codec. We further show that our proposed method has the potential to achieve greater improvements when coding point cloud videos.

\section{Neural Field based Compression}
\label{sec:form}




\subsection{Represent Point Cloud with Neural Field}

Neural fields, \textit{i.e.} neural networks overfitted on a given scene, are capable of constructing implicit volumetric functions to represent 3D environments~\cite{mildenhall2020nerf}. To utilize neural fields for point cloud compression, three problems need to be addressed: 1) We need to represent a point cloud by a volumetric function, so a network can be trained to fit the function and reconstruct a 3D volume. 2) the entropy of the neural field parameters should be controllable to reach different bit-rates.  3) With the original NeRF structure, we need to query every voxel to reconstruct the occupancy, where a large portions of them are empty. Such process is time consuming and not suitable for decoding. Directly converting a point cloud to the volumetric form, \textit{i.e.} voxel grid, is also memory inefficient. Besides, as we need many parameters to fully characterize the entire voxel grid, training the network and controlling the bit-rate is difficult.

We address these issues by constraining the volumetric space represented by the neural field, and using a convolutional network to generate groups of points at one time. We first build a shallow octree from the point cloud. Such a shallow octree only takes very few bits to represent. Each octree leaf node that is 1 corresponds to a non-empty  subregion (thereafter called a cube) of the original point cloud space that contains points.  We use a neural network along with a input latent code to represent the occupancy in each non-empty cube, so the burden to characterize the entire volume is lightened.  A shared network that can reconstruct all cubes of a point cloud, along with their respective latent codes, are learned to represent all the cubes. Unlike the original NeRF work, where the input is a user-specified view point, and only the network is trained to generate the 2D projection from that view point, we learn both the network and the latents associated with all the cubes. We call the collection of the network parameters and the latents as the neural field.

The framework of the proposed approach is illustrated in Fig.~\ref{fig:frame}. 
Assuming the original point cloud is voxelized and represented with an $M+N$ level octree. We take the first $M$ levels as the shallow octree $T$.
It describes the point cloud $\mathbf{x}$  at a coarse resolution.
Each non-empty leaf node of $T$ is associated with a subtree of level $N$, corresponding to a $(2^N,2^N,2^N)$ binary cube. For example, if the original point cloud is described by a 10-level octree, we may choose $M$=5, and $N$=5, so that each cube has a shape of $(32, 32, 32)$.
The $k$-th cube is associated with a trainable latent code ${z}_k$. 

Given a point cloud, the encoder learns the network parameters $\mathbf{y}$ and the latent codes $\mathbf{z}=\{ z_k, \forall k\}$ through backpropagation using a rate-distortion loss function. The quantized $\mathbf{y}$ and $\mathbf{z}$ are entropy coded, and the resulting bits together with the bits describing the shallow tree $T$ form the coded representation of the original point cloud. At the decoder,  we reconstruct the voxelized representation of the point cloud by feeding the latent code ${z}_k$ through a shared neural network with parameters $\mathbf{y}$ to reconstruct each non-empty cube.   


\subsection{Neural Field Network Structure}


In this section, we describe the neural network structure for generating a non-empty cube in more detail. Because we want to generate a size $(2^N,2^N,2^N)$ binary cube, and we let the latent code  ${z}_k$ to have a spatial dimension of $(2^L,2^L,2^L), L<N$ with $J$ channels. The generator consists of several convolution layers, some with transposed convolutions as the upsampling method. 
We add a {\it latent code generator} in front of the {\it cube generator} consisting of a $1\times 1$ convolutional layer and a 3D GDN~\cite{balle2015density}, followed by a rounding function. The input to the {\it latent code generator} $v_k$ has the same dimension as  $z_k$.  The purpose of the $1\times 1$ convolutional layer and the 3D GDN layer is to decorrelate the elements in $z_k$ and furthermore make each element (before rounding) have a distribution similar to Gaussian~\cite{balle2015density}.  An example network with $N=5, L=1, J=4$ is shown in Fig.~\ref{fig:structure}. Notice that this is a very light neural network, thanks to the small size of the cube.


\subsection{Rate Constraint Loss}

We will optimize the latent codes and the network parameters by minimizing a rate-distortion loss function,
\begin{equation}
    \centering
    \small
\begin{split}
    \mathscr{L} = \mathscr{L}_{R_z} + \mathscr{L}_{R_y}  + \lambda \mathscr{L}_D.
\end{split}
\end{equation}
Different $\lambda$ is set to achieve different R-D tradeoff. 
We first discuss how to estimate the rate terms $\mathscr{L}_{R_z} $ and $\mathscr{L}_{R_y}$
in the following.

By assuming each element $z_i$ in $\mathbf z$ (before quantization) follows a Gaussian distribution with mean $\mu_z$ and scale $\sigma_z$, we can calculate the lower bound of the bits needed to encode the quantized $\mathbf z$, as,
\begin{equation}
\small
    \mathscr{L}_{R_z} = \sum_{i} -\log(q(z_i)),
\end{equation}
where $q(z_i)$ is the estimated probability of $z_i$, given by,
\begin{equation}
\small
    q(z_i) = \phi(\frac{z_i-\mu_z+\Delta/2}{\sigma_z}) - \phi(\frac{z_i-\mu_z-\Delta/2}{\sigma_z}),
    \label{eq:p}
\end{equation}
where $\Delta$ equals  the quantization step, and $\phi$ is the standard normal cumulative distribution function.
We assume each element of the network parameter $\mathbf y$ before quantization also has a Gaussian distribution, with parameters $\mu_y, \sigma_y$, from which we can estimate the lower bound on the rate for encoding $\mathbf y$, denoted by $\mathscr{L}_{R_y}$. We use a quantization step size of $\Delta=1/16$ for the network parameters and $\Delta=1$ for the latent code. During training, to  estimate the bit-rate for  $\mathbf{z}$ and $\mathbf{y}$ and make the decoder quantization resilient,  we add a uniform noise $\mathcal{U}(-\Delta/2, \Delta/2)$ to each element in $\mathbf z$ and $\mathbf{y}$ rather then performing quantization. 


The encoding of a point cloud is de facto the training process of the network using backpropagation, but we update both network parameters $\mathbf y$ and the input $v_k, \forall k$ (and consequently the latent code ${ z}_k, \forall k$). In addition, we also update the distribution parameters $\mathbf{q}=\{ \mu_z, \sigma_z, \mu_y, \sigma_y \}$. Since $\mathbf{q}$ only contains a few floating point values, after the rate-distortion training, we simply transmit $\mathbf{q}$ in its original 32-bit floating point form to the decoder.
Then the quantized latent code $\mathbf{z}$ for all leaf nodes and the quantized network parameters $\mathbf{y}$ are entropy coded based on $\mathbf{q}$.

Note that the parameters in the {\it latent code generator} part of the neural network (see Fig.\ref{fig:structure}) will be trained together with the {\it cube generator} part, but they do not  need to be encoded, as the decoder only need to apply the latent code to the {\it cube generator} part to decode each cube. Therefore $\mathbf{y}$ only includes parameters in the {\it cube generator}. Similarly, during training, we update $v_k$ instead of $z_k$. We encode the final optimal $z_k$ corresponding to the optimized $v_k$. 

\subsection{Network Parameters Initialization}
\begin{figure}[t]
    \centering
    \includegraphics[width=0.7\linewidth]{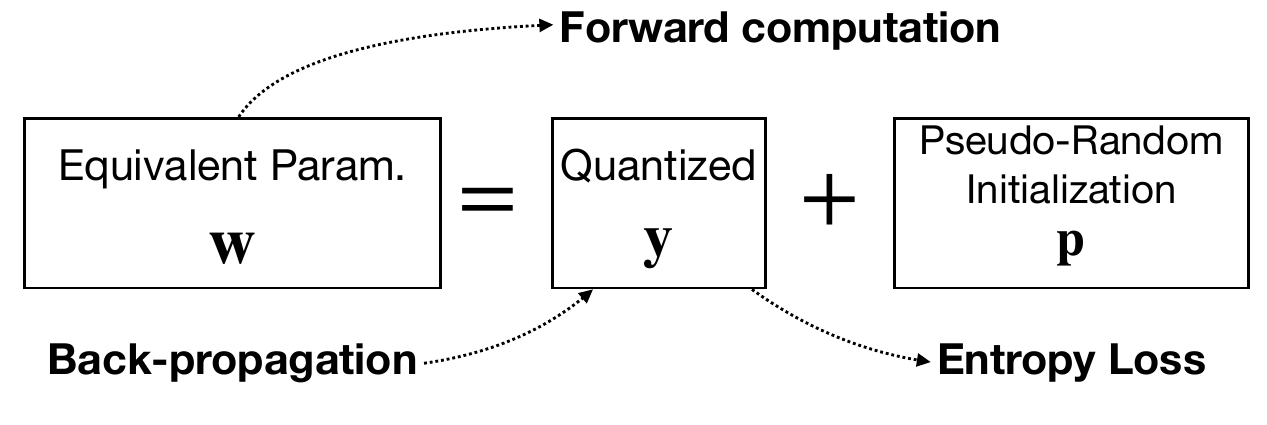}
    \caption{Separation of initialization and the coded network parameters.}
    \label{fig:zero}
\end{figure}

Since we need to entropy code both the network parameters and the latent representation, we need to apply rate constraints on both during the training. In this circumstance, the commonly used random weight initialization introduces high entropy to the parameters at the beginning, making it hard to control the bit-rate throughout the training process. To address this problem, we propose to separate the initialization from the coded network parameters, as illustrated in Fig.~\ref{fig:zero}. 
%
We use a Kaiming pseudo ramdon initialization~\cite{he2015delving} tensor $\mathbf{p}$, with the same shape as $\mathbf{y}$, to initialize the network parameters. The initialization $\mathbf{p}$ is fixed for all point clouds and shared with the decoder. The actual network parameters are represented by  $\mathbf{w}=\mathbf{p}+\mathbf{y}$.  During the training, we only update $\mathbf y$. 
 Because  $\frac{\partial \mathscr{L}}{\partial \mathbf{w}} = \frac{\partial \mathscr{L}}{\partial \mathbf{y}}$,  we can use the standard backpropagation gradient to update $\mathbf y$. 
By initializing  $\mathbf{y}$ with zeros, we can limit its entropy throughout the entire training process.
For the latent code, we initialize with zeros for all elements, which help to minimize the rate of the optimized latent code.


\subsection{Distortion Loss Function}
\label{sec:distortion}
The distortion loss function we use is based on the binary focal loss, formulated as,
\begin{equation}
\centering
\begin{split}
    &\mathscr{L}_{\text{focal}} = - \sum_{i} \alpha_i (1-F_i)^\gamma \log F_i, \\
    & F_i = \begin{cases}
      q_i &\text{ if } y_i=1 \\
      1-q_i &\text{ if } y_i=0
    \end{cases},  \alpha_i = \begin{cases}
      \alpha &\text{ if } y_i=1 \\
      1-\alpha &\text{ if } y_i=0
    \end{cases},
\end{split}
\label{eq:focal}
\end{equation}
where $y_i \in \{0,1\}$ is the ground truth occupancy of voxel $i$, and $q_i$ is the predicted probability that voxel $i$ is occupied. Since in the voxel grid generated from a typical point cloud, the non-empty voxels only occupy a very small portion, the original BCE loss tends to make the trained network predict all zeros. The focal loss is used to balance the empty and non-empty voxels during training, which applies a higher weight to the voxels that are non-empty. In our experiments, $\alpha$ is set to the portion of empty voxels in the binary grids. 

However, the focal loss is designed for classification problems, where every sample with the same ground truth label has the same weight. In voxel grid prediction, empty voxels far away from the surface, if misclassified as occupied, introduce greater geometric distortion. Hence, it is intuitive to apply a higher penalty to these far-away errors. Following this idea, in this work, we propose the distance-weighted focal loss function, formulated as,
\begin{equation}
    \centering
    \small
\begin{split}
    & \mathscr{L}_{\text{d}} = - \sum_{i} \alpha_i(1-F_i)^\gamma D_i \log F_i, \text{ } D_i = \min_{\mathbf{p}\in S} ||\mathbf{p}_i - \mathbf{p}||_2,
\end{split}
\label{eq:dweighted}
\end{equation}
where $\mathbf{p}_i \in \mathbb{R}^3$ is the 3D coordinate of voxel $i$ and the minimization is taken over the set $S$ of all voxels that are occupied in the ground truth point cloud. 
For positions farther away from any occupied voxel in the ground truth, the loss function applies higher penalties on false positive predictions. 
In our implementation, before training, we compute the distances from every possible voxel  to its nearest point in the original point cloud in advance. Hence no extra computation of distances is needed during the training.

\begin{figure*}[t]
    \centering
    \begin{subfigure}[h]{0.22\linewidth}
    \includegraphics[width=1\linewidth]{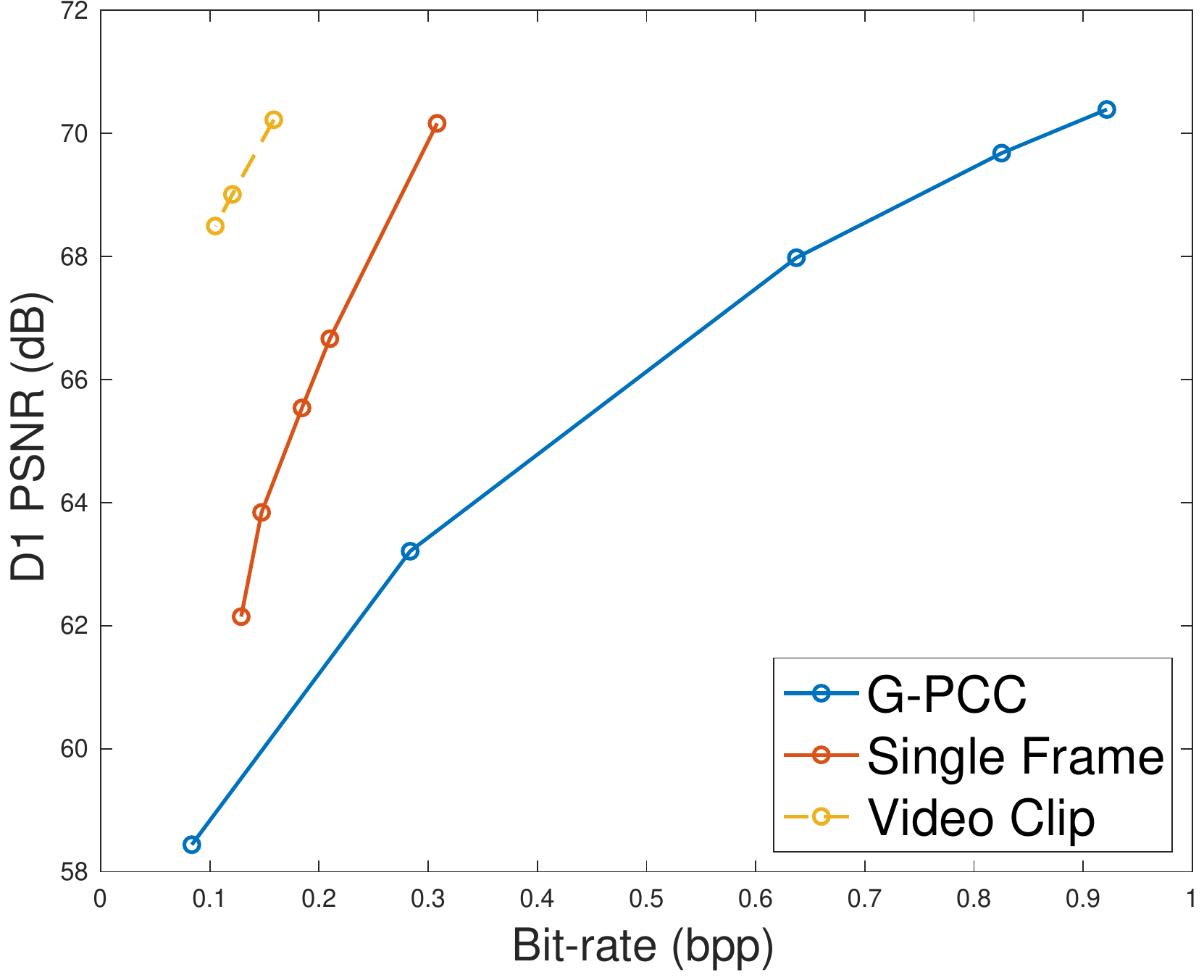}
    \caption{{longdress\_vox10\_1300}}
    \end{subfigure}
    \begin{subfigure}[h]{0.22\linewidth}
    \includegraphics[width=1\linewidth]{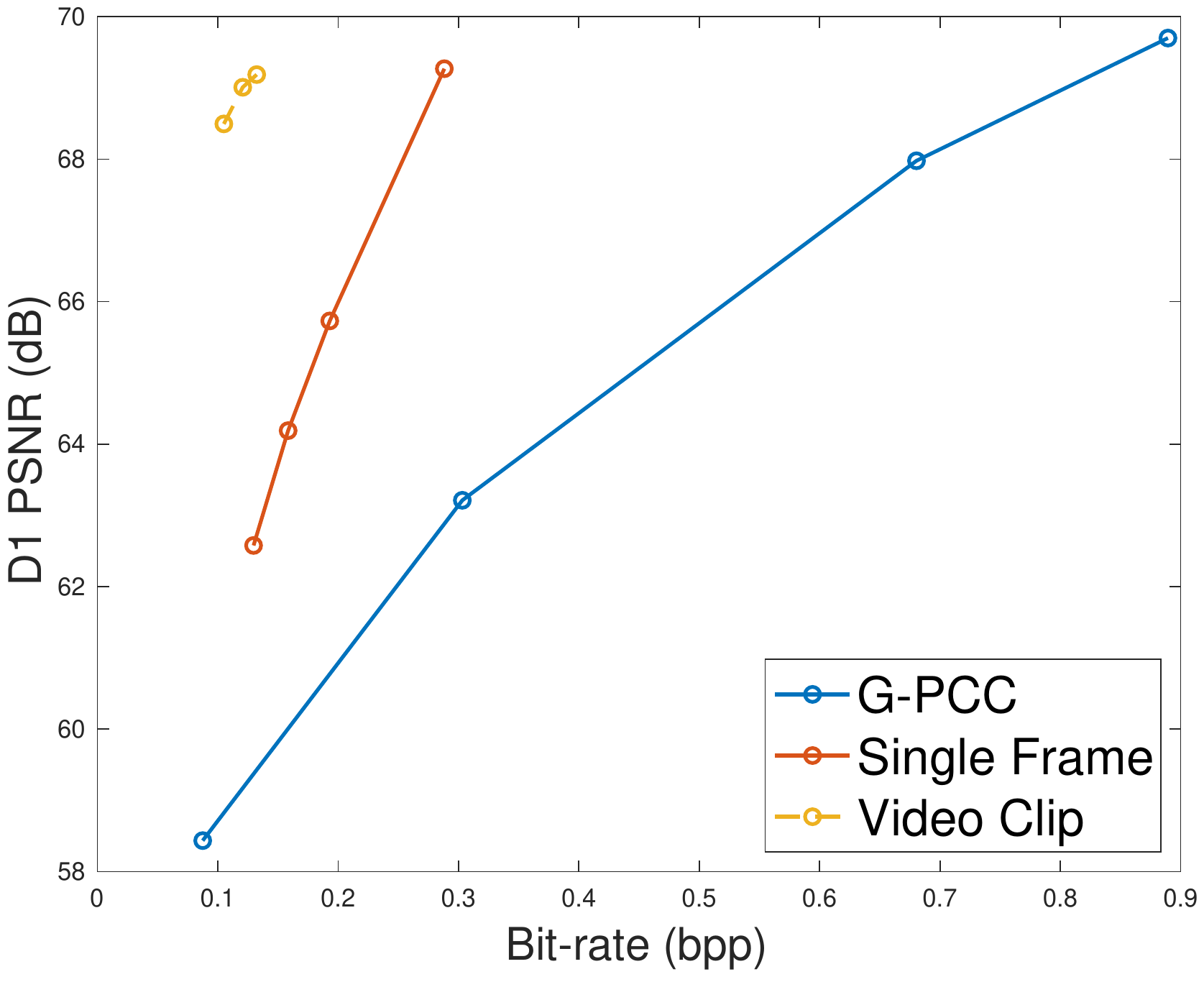}
    \caption{{redandblack\_vox10\_1450}}
    \end{subfigure}
    \begin{subfigure}[h]{0.22\linewidth}
    \includegraphics[width=1\linewidth]{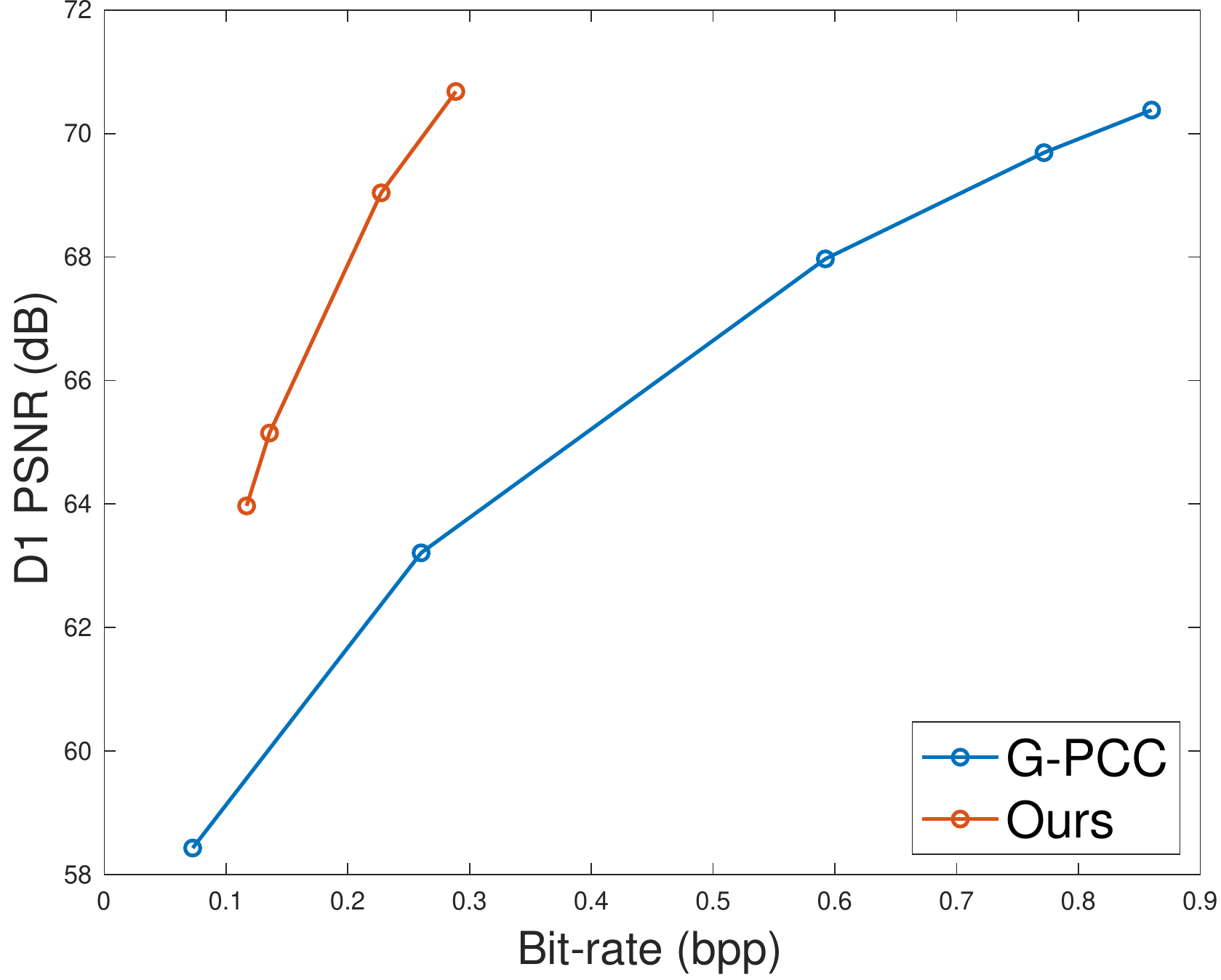}
    \caption{{loot\_vox10\_1000}}
    \end{subfigure}
    \begin{subfigure}[h]{0.22\linewidth}
    \includegraphics[width=1\linewidth]{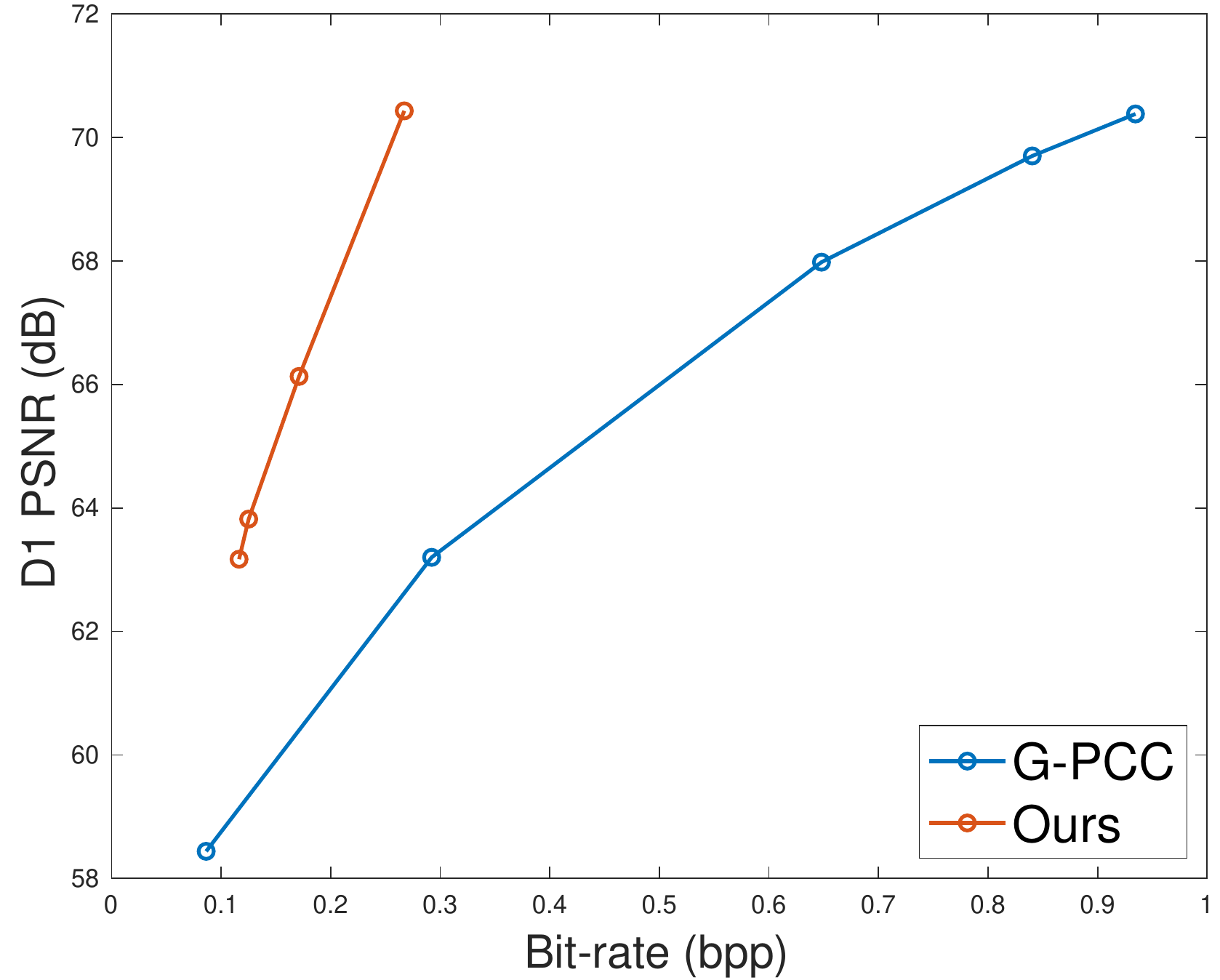}
    \caption{{soldier\_vox10\_0540}}
    \end{subfigure}
    \caption{R-D curves by the proposed method and G-PCC.}
    
\vspace{-6mm}
    \label{fig:rd8i}
\end{figure*}


Since we have multiple upsampling layers in the generator network, we can have a better supervision by calculating distortion loss at multiple spatial scales, \textit{i.e.} $(2^{N},2^{N},2^{N})$, $(2^{N-1},2^{N-1},2^{N-1})$ and $(2^{N-2},2^{N-2},2^{N-2})$ as shown in Fig.~\ref{fig:structure}. The distortion term is the summation of the distortion loss at different scales, as,
\begin{equation}
    \centering
\begin{split}
    & \mathscr{L}_D = \mathscr{L}_{D_1} + \mathscr{L}_{D_2} + \mathscr{L}_{D_3},
\end{split}
\end{equation}
where $\mathscr{L}_{D_2}$ and $\mathscr{L}_{D_3}$ are focal losses calculated with the down-sampled ground truth voxel grids. For $\mathscr{L}_{D_1}$ we use the distance weighted focal loss in Eq.~(\ref{eq:dweighted}). 


\subsection{Coding Procedure}
Fig.~\ref{fig:frame}b summarizes the encoding and decoding processes. As shown, two bit-streams are formed by the encoder to represent a point cloud, \textit{i.e.}, the shallow octree $T$  and the neural field $(\mathbf{y}, \mathbf{z})$. %
Since the shallow subtree $T$ has only a small number of levels,  in our experiments we code them by simply traversing the octree in BFS order and write out every occupancy bit, which account for a very small portion of the total bits.
We assume that the parameters in $\mathbf{y}$ and/or $\mathbf{z}$ have Gaussian distribution, with Gaussian parameters described by $\mathbf{q}$. The entropy coding will be aided with $\mathbf{q}$. In Fig.~\ref{fig:frame}b, we use $W$ to represent the bits for $\mathbf{y}$ and $\mathbf{z}$. In general, $\mathbf{q}$ is content-dependent and will also need to be coded. As described earlier, since $\mathbf{q}$ only contains a few floating point values, we simply use the original 32-bit floating point form to represent each value. The bits for $T, q$, $\mathbf{y}$, and $\mathbf{z}$ form the bit-stream for $x$.
%
%
At the  decoder, we  first entropy decode $\mathbf{y}$ and $\mathbf{z}$ based on $\mathbf{q}$, and construct the network using $\mathbf{y}$.  We then enumerate all the non-empty leaf nodes of $T$ and use the decoded latent for each leaf node as the input to the network. The network generates  occupancy probabilities for all voxels in each cube. The probabilities  are binarized to reconstruct the point cloud.

\section{Experimental Results}
\subsection{Experimental Settings}

We conduct experiments on the 8i Voxelized Full Bodies (8iVFB) dataset (\textit{Longdress}, \textit{Loot}, \textit{Redandblack}, \textit{Soldier}), which are adopted by MPEG Common Testing Condition~\cite{ctc}. The point clouds are all of bit-depth 10. We choose $M = 5$ and $N=5$. We use a spatial dimension of $(2,2,2)$ (i.e. $L=1$) for the latent code for each leaf node.   We compare our method with G-PCC TMC13~\cite{gpcc}, for which the octree coding scheme is used. Since we generate a probability for each cube element, we threshold the probabilities to  produce the reconstructed point cloud.  In the experiment, we choose the threshold that balance the two PSNRs in determining the D1 PSNR \cite{tian2017geometric} for each point cloud. The threshold is signaled to the decoder with 32 bits.

We evaluate our proposed method in terms of R-D performance. We measure the bit-rate as bit-per-point (bpp), where the number of bits is the summation of the number needed to code the first $M$-level octree, the network parameters, the latent representations, and the distribution parameters. The distortion is measured with the point-to-point error PSNR (\textit{a.k.a.} D1 PSNR).
\subsection{R-D Performance}

The comparison in R-D performance of the proposed method and G-PCC on single frames of the point clouds is shown in Fig.~\ref{fig:rd8i}. For the same point cloud, we reach different R-D points by training multiple networks with different $\lambda$, different number of channels (i.e. $J$) in the latent code  and different widths of the generator network (\textit{i.e.} the number of output channels for the intermediate layers). The network settings are fixed for the same $\lambda$ among the entire testing dataset.
As shown, on all four point clouds, our method achieve improved R-D performance than G-PCC. 

The proposed method can be directly applied to dynamic point clouds. Since frames in a dynamic point cloud sequence  share common geometry patterns, we can share the network parameters among all leaf nodes in all frames from a dynamic point cloud sequence. The bits to encode the network parameters are therefore amortized over the larger number of points. We further conduct the dynamic point cloud compression experiments on the \textit{Longdress} and \textit{Redandblack} sequences, where we code  16 successive frames in each sequence, \textit{i.e.} \textit{longdress\_vox10\_1300} to \textit{longdress\_vox10\_1315} in \textit{Longdress}, \textit{redandblack\_vox10\_1450} to \textit{redandblack\_vox10\_1465} in \textit{Redandblack}.
Since G-PCC only supports all-intra coding mode, we can take the G-PCC R-D curve as a reference to the dynamic coding scenario. As shown in Fig.~\ref{fig:rd8i} (a, b), dynamic point cloud can significantly benefits from our method, where the neural field learns the mutually shared patterns among leaf nodes over 16 frames.
 
\subsection{Ablation Study on Architecture and Loss Function}

\begin{figure}[t]
    \centering
    \includegraphics[width=0.7\linewidth]{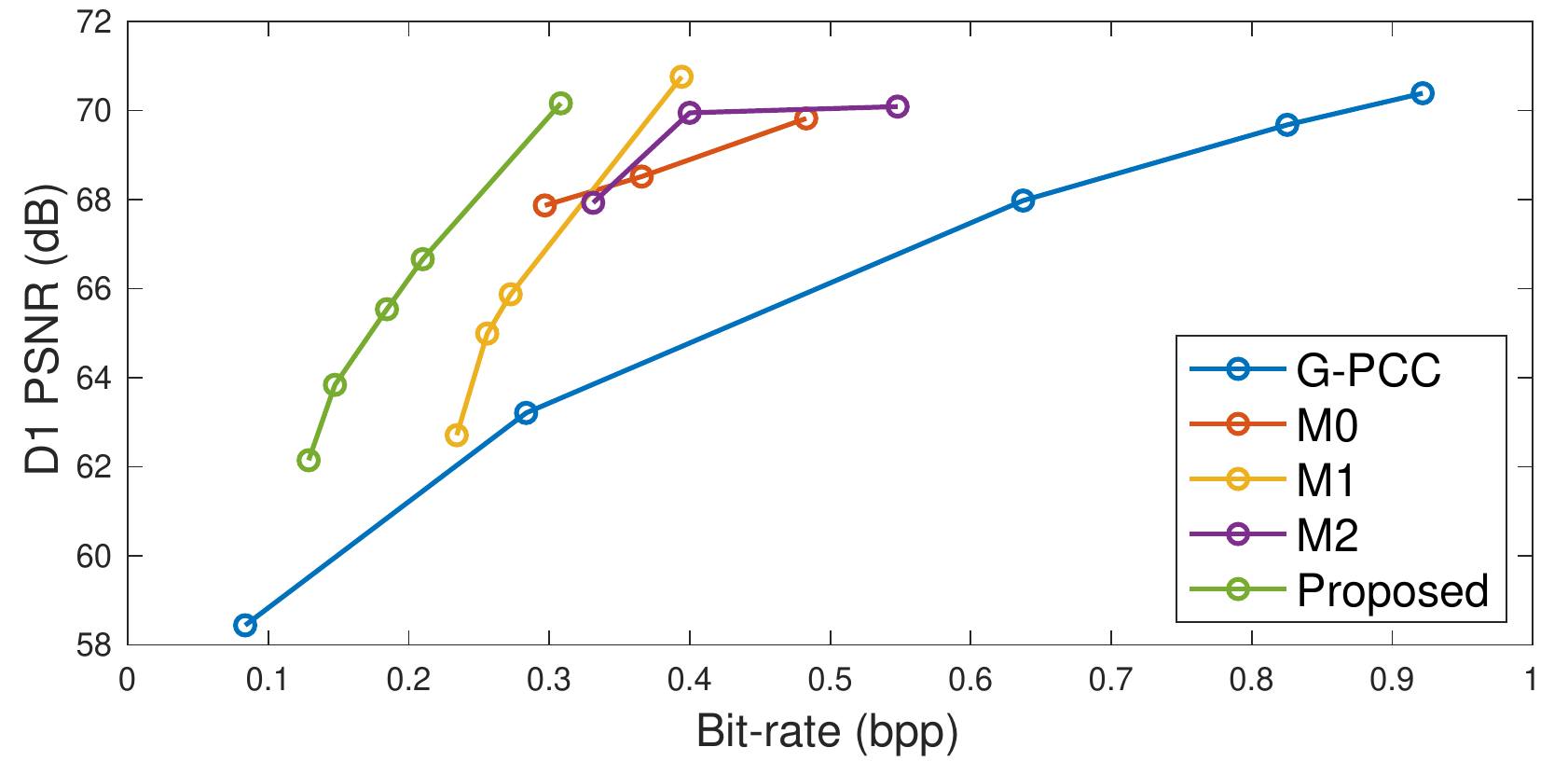}
    \caption{Rate-distortion performance evaluation for multiple variations of the proposed method and G-PCC. 
    }
    \label{fig:rd}
    \vspace{-2mm}
\end{figure}


To investigate the contribution by different components of the proposed scheme, we compare different models obtained with different configurations in a ablation study. The detailed configurations of these models are shown in Table~\ref{tab:options}. If \textit{D-weighted Loss} is not checked, the original focal loss is used. If \textit{Rate Loss} is not checked, the bit-rate term is removed from the loss function. 
If \textit{Init. Separation} is not checked, the actual network parameters are coded, rather than the difference from the Kaiming initialization. 
\begin{table}[t]
    \centering 
    \caption{Model variations in the ablation study. }
    \begin{tabular}{c|c|c|c}
    \hline
        Variation & D-weighted Loss & Rate Loss & Init. Separation \\
        \hline
        M0 & \checkmark & & \checkmark  \\ 
        M1 & & \checkmark & \checkmark  \\
        M2 & & \checkmark&  \\
        Proposed & \checkmark & \checkmark & \checkmark  \\
        \hline
    \end{tabular}
    \label{tab:options}
    \vspace{-3mm}
\end{table}

 
 
 
 
The results are shown in Fig.~\ref{fig:rd}. Comparing M0 to Proposed, we observe that without the rate-distortion training, the bit-rates are much higher at the same distortion level, and the lower bit-rate range cannot be reached. The improvement of M1 over M2 demonstrates that initialization separation enables the network to be extended to low-rate range. Finally, comparing M1 and Proposed, we see that the distance-weighted focal loss helps improve the rate-distortion performance significantly and consistently over the entire range.

\section{Conclusion}
In this paper, we introduce a novel approach to point cloud compression.  To overcome the challenges in designing transforms and motion-compensated prediction techniques for point clouds, we utilize an iterative optimization process as the way to find a good compressive representation for a point cloud. Specifically, we divide the entire space into small cubes and represent each non-empty cube by a neural network and an input latent code. The network is shared among all the cubes in a single frame or multiple frames, to exploit the spatial and temporal redundancy. We train the network parameters and latent codes as the representation for all the non-empty cubes of a point cloud. To the best of our knowledge, this is the first work to utilize learned neural fields for point cloud compression. We develop a series of techniques to control the bit-rates and improve the reconstruction quality. Experimental results show that these techniques are effective, and our method achieves a better rate-distortion performance than G-PCC.  Even though the R-D performance of the proposed approach for static point clouds are below some of the recently published works e.g.,\cite{wang2021lossy,wang2021multiscale},  it opens a new avenue for exploration, and future research in this direction may significantly improve the performance. More importantly, such an approach has great potential to effectively compress point cloud videos, because it can easily exploit the inter-frame redundancy through training a shared network for multiple frames. 
\vspace{-1mm}
\section*{Acknowledgement}
This work was supported in part by an unrestricted gift from FutureWei Technologies, Inc. to support fundamental research.
\vspace{-2mm}
\bibliographystyle{IEEE_short}
\bibliography{egbib_short}

\end{document}